\documentclass[sigconf]{acmart}

\usepackage{algorithm}
\usepackage{multirow}
\usepackage{amsmath}

\usepackage{amssymb}
\usepackage{amsthm}
\usepackage{graphicx}
\usepackage{enumitem}

\theoremstyle{definition}

\usepackage{algorithm}
\usepackage[noend]{algpseudocode}
\usepackage{bm}
\usepackage{makecell}
\usepackage{tikz}
\usetikzlibrary{positioning, arrows.meta, shapes.geometric, calc}

\tikzset{
  >=Latex,
  box/.style={
    draw, rounded corners,
    align=center,
    inner sep=3pt, outer sep=0pt,
    minimum width=6.0cm, minimum height=6mm
  },
  boxJ/.style={box, fill=gray!20},
  boxG/.style={box, fill=green!20},
  boxB/.style={box, fill=blue!20},
  edge/.style={->, line width=0.9pt},
  edgeDashed/.style={edge, dashed},
  lab/.style={font=\scriptsize, fill=white, fill opacity=0, text opacity=1, inner sep=1pt}
}

\renewcommand\footnotetextcopyrightpermission[1]{}

\author{Jinzhou Tan}
\affiliation{%
  \institution{University of California, San Diego}
  \country{USA} 
}

\author{Gabriel Adineera}
\affiliation{%
  \institution{Texas A\&M University-Commerce}
  \country{USA} 
}

\author{Jinoh Kim}
\authornote{Corresponding author.} 
\affiliation{%
  \institution{Texas A\&M University-Commerce}
  \country{USA} 
}
\begin{document}
\title{ProgAgent: A Continual RL Agent with Progress-Aware Rewards}

\begin{abstract}
We present ProgAgent, a continual reinforcement learning (CRL) agent that unifies progress-aware reward learning with a high-throughput, JAX-native system architecture. Lifelong robotic learning grapples with catastrophic forgetting and the high cost of reward specification. ProgAgent tackles these by deriving dense, shaped rewards from unlabeled expert videos through a perceptual model that estimates task progress across initial, current, and goal observations. We theoretically interpret this as a learned state-potential function, delivering robust guidance in line with expert behaviors. To maintain stability amid online exploration—where novel, out-of-distribution states arise—we incorporate an adversarial push-back refinement that regularizes the reward model, curbing overconfident predictions on non-expert trajectories and countering distribution shift.By embedding this reward mechanism into a JIT-compiled loop, ProgAgent supports massively parallel rollouts and fully differentiable updates, rendering a sophisticated unified objective feasible: it merges PPO with coreset replay and synaptic intelligence for an enhanced stability-plasticity balance. Evaluations on ContinualBench and Meta-World benchmarks highlight ProgAgent's advantages: it markedly reduces forgetting, boosts learning speed, and outperforms key baselines in visual reward learning (e.g., Rank2Reward, TCN) and continual learning (e.g., Coreset, SI)—surpassing even an idealized perfect memory agent. Real-robot trials further validate its ability to acquire complex manipulation skills from noisy, few-shot human demonstrations.

\end{abstract}
\keywords{Machine Unlearning, Large Language Models, Contrastive Learning, Representation Reshaping, Concept Decoupling, AI Safety}

\maketitle
\section{INTRODUCTION}

Endowing robots with the ability to learn continually—acquiring new skills while retaining and refining past knowledge—represents a central ambition in robotics. This paradigm of \textit{lifelong reinforcement learning (RL)} sharply contrasts with conventional RL\cite{pan2025surveycontinualreinforcementlearning,khetarpal2022continualreinforcementlearningreview,KABB,GAM,MAB,MAT}, which typically assumes a static environment and trains policies from scratch\cite{DBLP:journals/corr/abs-1708-05866}. In real-world scenarios, however, robots must navigate evolving task sequences, shifting objectives, and unstructured sensory inputs. These dynamic conditions expose persistent bottlenecks: first, \textbf{catastrophic forgetting}, where adapting to new tasks overwrites prior capabilities, undermining long-term autonomy \cite{DBLP:journals/corr/abs-2012-13490,MCCLOSKEY1989109,CF,3DAgent,z22}; second, the \textbf{reward specification problem}, as crafting dense, well-shaped rewards for varied manipulation tasks demands intensive labor and often proves impractical, constraining scalability outside controlled benchmarks.

Building on this foundation, prior research has tackled these issues through two main, yet largely disconnected, avenues. In continual RL, algorithmic strategies like parameter regularization, rehearsal buffers, and synaptic importance metrics have emerged to curb forgetting by preserving key knowledge from past tasks\cite{DBLP:journals/corr/abs-1802-07569,HTC,KAF,DrDiff,OSC}. Concurrently, advances in visual reward learning enable agents to derive rewards from unlabeled expert videos, sidestepping manual design. Despite these strides, the fields remain orthogonal: continual learning algorithms frequently overlook system-level optimizations for scalable training—such as just-in-time (JIT) compilation and parallelized rollouts—while reward models often falter under distributional shifts, as online exploration leads to non-expert states diverging from training data\cite{hu2018overcoming,DBLP:journals/corr/abs-1011-0686,MMCOT,VLMDONG,RAN,POT}. This separation creates a critical gap, hindering the development of unified agents that integrate \textbf{perceptual reward learning} with \textbf{scalable architectures for continual adaptation}.

To bridge this gap, we introduce \textbf{ProgAgent}, a continual RL agent that fundamentally unifies \emph{progress-aware reward estimation} with a \emph{JAX-native architecture}. Algorithmically, ProgAgent trains a perceptual model to predict task progress from initial, current, and goal observations, generating dense, shaped rewards without action labels. This design draws theoretical motivation from potential-based shaping: assuming expert demonstrations exhibit monotonic progress toward goals, the predictions form an implicit potential function that aligns exploration with expert trajectories. Recognizing the challenges of online exploration, we further incorporate an adversarial \textit{push-back} refinement to regularize rewards, suppressing overconfidence on off-distribution states and enhancing robustness. On the systems side, ProgAgent embeds this reward model within a fully differentiable, JIT-compiled pipeline. By transforming the entire training loop—including data collection, reward updates, and policy optimization—into highly optimized kernels via JAX's JIT compilation, it enables massively parallel rollouts across thousands of environments for rapid data generation, minimizes gradient variance through batched computations for more stable updates, and supports seamless scalability to handle complex, multi-task sequences without prohibitive computational overhead—directly addressing the algorithm-system divide that has limited prior continual RL efforts, where advanced algorithms were often too resource-intensive to implement efficiently at scale.

These innovations translate to substantial empirical gains, as validated through rigorous evaluations across diverse settings. On the \textit{ContinualBench} benchmark, ProgAgent markedly reduces forgetting, delivering state-of-the-art average performance and regret metrics. In six demanding \textit{Meta-World}\cite{DBLP:journals/corr/abs-1910-10897} manipulation tasks, it surpasses leading visual reward approaches like Rank2Reward, exhibiting exceptional sample efficiency and adaptability. Extending to the physical domain, real-robot experiments affirm ProgAgent's practicality: it effectively learns complex manipulation skills from a handful of noisy human demonstrations, succeeding even when half the data involves failures.

In summary, this work advances lifelong robotic learning through three key contributions:
\begin{enumerate}
    \item \textbf{A progress-aware reward model} that extracts dense, shaped signals from unlabeled expert videos, theoretically grounded as a state-potential function for expert-aligned guidance. This approach bypasses the need for action labels and provides a monotonic progress signal that accelerates policy optimization while preserving optimality guarantees.
    \item \textbf{An adversarial refinement mechanism} that stabilizes the reward model by countering overconfident predictions on non-expert trajectories, ensuring reliable performance amid distribution shifts. By regularizing predictions on exploratory data, it prevents misleading rewards that could derail learning, making the model robust for continual, online adaptation.
    \item \textbf{A unified JAX-native architecture} that JIT-compiles the full reward and policy optimization loop, enabling high-throughput parallelization and integration with advanced continual learning techniques for superior scalability. This design not only boosts computational efficiency but also allows for reproducible, large-scale experiments, bridging the gap between algorithmic innovation and practical system deployment.
\end{enumerate}


\section{RELATED WORKS}

Our work integrates three traditionally distinct research areas: continual reinforcement learning, reward learning from perception, and high-throughput RL systems. As autonomous agents are increasingly deployed in complex, open-world environments, the demand for scalable, self-sustaining learning paradigms has intensified. By synthesizing insights from these domains, ProgAgent creates a cohesive framework that addresses their respective limitations holistically. Below, we review key contributions in each area and highlight how our approach builds upon and unifies them.

\subsection{Continual Reinforcement Learning}

The core challenge in continual reinforcement learning (CRL) is mitigating \textbf{catastrophic forgetting}, wherein policies abruptly overwrite prior knowledge during adaptation to new tasks. This issue is particularly exacerbated in RL compared to supervised learning due to the inherent non-stationarity of the agent's observation distribution and the shifting nature of optimal behavior policies. Existing methods generally fall into three categories. \textbf{Regularization-based} techniques penalize modifications to parameters critical for previous tasks; prominent examples include Elastic Weight Consolidation (EWC), which employs a Fisher information matrix to gauge parameter importance \cite{DBLP:journals/corr/KirkpatrickPRVD16}, and Synaptic Intelligence (SI), which computes importance dynamically during the training trajectory \cite{DBLP:journals/corr/ZenkePG17,DBLP:journals/corr/abs-1711-09601,SGN,z15,z14,z13,M2,M3,M4}. \textbf{Rehearsal-based} strategies preserve past experiences by storing and replaying data, ranging from basic experience replay \cite{DBLP:journals/corr/abs-1811-11682} to advanced coreset selection or distillation techniques that curate compact, representative buffers to prevent memory explosion \cite{DBLP:journals/corr/abs-1902-10486,DBLP:journals/corr/ShinLKK17,z16,z20,M5,M6}. \textbf{Architecture-based} approaches, such as Progressive Neural Networks \cite{DBLP:journals/corr/RusuRDSKKPH16}, expand model capacity dynamically to isolate new tasks without interfering with old ones \cite{DBLP:journals/corr/abs-1711-05769,DBLP:journals/corr/abs-1904-00310}.

ProgAgent employs a hybrid strategy, intelligently combining SI regularization with coreset replay. Unlike prior efforts that deploy these mechanisms independently—often leading to suboptimal trade-offs between stability and plasticity, or incurring prohibitive computational costs—ProgAgent fuses them into a single, unified objective. This integration is made highly practical through our high-throughput architecture, which supports efficient, large-batch computation and enables a more balanced handling of knowledge retention and adaptation across lengthy task sequences \cite{DBLP:journals/corr/abs-1802-07569}.

\subsection{Learning Rewards from Perception}

Hand-crafting dense rewards poses a significant bottleneck in robotics. In complex manipulation or locomotion tasks, defining a reward function that perfectly aligns with human intent is notoriously difficult and highly susceptible to reward hacking, prompting extensive research into perceptual reward learning from raw data like unlabeled videos. Early imitation learning methods, including Behavioral Cloning (BC) and Inverse Reinforcement Learning (IRL), typically demand substantial expert action annotations or costly online environment interactions \cite{6796843, 10.5555/645529.657801}. Modern techniques circumvent these stringent requirements: \textbf{contrastive methods} derive rewards by enhancing representation similarities between proximate states while diminishing those for temporally or visually distant ones \cite{DBLP:journals/corr/SermanetLHL17}, and \textbf{ranking-based methods} utilize temporal ordering or human feedback, as seen in Rank2Reward's learned ranker \cite{yang2024rank2rewardlearningshapedreward} or Temporal Cycle Consistency (TCC) for self-supervised video representations \cite{DBLP:journals/corr/abs-1904-07846, christiano2023deepreinforcementlearninghuman}.

ProgAgent aligns most closely with progress-modeling paradigms, such as Temporal Difference Networks (TCN) for temporal distance prediction and goal-conditioned progress estimation \cite{wang2025divergenceaugmentedpolicyoptimization}. We extend this paradigm by conceptualizing progress strictly as a learned potential function, which provides rigorous theoretical guarantees for policy invariance and shaped exploration. Crucially, to overcome a common shortfall in purely unsupervised reward models—susceptibility to severe distribution shifts and false positives from visually familiar but unexplored states—we introduce an adversarial push-back refinement. This mechanism enhances robustness during online learning, ensuring the agent remains on a functionally valid trajectory and bridging critical gaps in prior models \cite{ebert2017selfsupervisedvisualplanningtemporal}.

\subsection{High-Throughput Reinforcement Learning Systems}

The intensive computational needs of RL have spurred the rapid development of accelerator-optimized simulation systems. Historically, RL systems suffered from severe CPU-GPU communication bottlenecks due to the disconnect between environment stepping and policy updates. Platforms like Isaac Gym \cite{DBLP:journals/corr/abs-2108-10470} and Brax \cite{DBLP:journals/corr/abs-2106-13281} eliminate this overhead and achieve dramatic speedups by executing thousands of parallel simulations directly on GPUs. They are often powered by just-in-time (JIT) compilation tools like JAX \cite{jax2018github}, which elegantly convert Python code into efficient XLA kernels \cite{6386109}. While these frameworks excel at accelerating single-task RL, they rarely address the multifaceted memory and computational demands of continual settings, such as sequential task adaptation, dynamic buffer management, and integrated reward refinement.

ProgAgent advances this computational landscape by tailoring a JAX-native architecture specifically to CRL's unique complexities. By encapsulating the simulator in pure functions and completely JIT-compiling the entire training loop—including environmental data collection, replay buffer sampling, reward model updates, and multi-objective policy optimization—we unlock the massive throughput required for a sophisticated, unified CRL algorithm. By eliminating host-device data transfers for complex regularization terms, this design not only makes advanced continual mechanisms computationally viable but also ensures unprecedented scalability, reproducibility, and time-efficiency in lifelong learning scenarios.
\section{METHOD}

Building on the foundational challenges and prior work outlined in the introduction and related literature, we present ProgAgent as a continual RL framework that fundamentally unifies progress-aware reward estimation with a JAX-native architecture. This integration directly addresses the algorithm-system divide highlighted earlier, enabling scalable training in lifelong learning scenarios. Unlike previous methods that treat reward learning and architectural optimization as separate concerns, ProgAgent fuses them into a cohesive, end-to-end pipeline. In this section, we formalize the problem setting, then elaborate on the algorithmic innovations for reward shaping and the system-level designs that ensure computational efficiency, aligning with the hybrid continual mechanisms and perceptual reward paradigms discussed in the related works.

\subsection{Problem Formulation}

We formalize the continual reinforcement learning (CRL) setting, where an agent encounters a sequence of tasks $\{\mathcal{T}_1, \mathcal{T}_2, \dots, \mathcal{T}_K\}$. Each task $\mathcal{T}_i$ is defined as a Markov Decision Process (MDP\cite{sutton2018reinforcement,kaelbling1996reinforcement,lin2025stormsearchguidedgenerativeworld}) $\langle \mathcal{S}, \mathcal{A}, P_i, r_i, \gamma \rangle$, sharing a common state space $\mathcal{S}$ and action space $\mathcal{A}$, but featuring potentially distinct transition dynamics $P_i$ and reward functions $r_i$. This setup captures the evolving nature of real-world robotic environments, where tasks build upon or diverge from previous ones, as emphasized in the introduction\cite{712192}.

The agent's policy $\pi_\theta(a|s)$ must optimize under dual constraints: mitigating \textbf{catastrophic forgetting} to sustain performance on prior tasks $\mathcal{T}_{1:i-1}$ while adapting to the current $\mathcal{T}_i$, and overcoming \textbf{reward specification} challenges by deriving dense, shaped signals without manual, task-specific designs. Formally, the goal is to maximize average performance across all seen tasks via parameters $\theta$:
\begin{equation}
\max_\theta \ \frac{1}{i}\sum_{j=1}^i \mathbb{E}_{\tau \sim \pi_\theta, P_j}\Big[ \sum_{t=0}^\infty \gamma^t r_j(s_t, a_t) \Big].
\label{eq:crl_objective}
\end{equation}
This objective, unlike standard single-task RL, explicitly accounts for multi-task retention, highlighting the need for mechanisms that balance plasticity and stability—a key gap in prior CRL approaches, as discussed in related works. By optimizing this averaged return, ProgAgent ensures long-term autonomy without sacrificing adaptability.

\begin{figure}[h!]
\centering
\begin{tikzpicture}[
  node distance=15mm,
  every node/.style={font=\small},
  >=Latex,
  box/.style={draw, rounded corners, align=center, inner sep=3pt, outer sep=0pt,
              minimum width=6.0cm, minimum height=6mm},
  boxJ/.style={box, fill=gray!20},
  boxG/.style={box, fill=green!20},
  boxB/.style={box, fill=blue!20},
  edge/.style={->, line width=0.9pt},
  edgeDashed/.style={edge, dashed},
  lab/.style={font=\scriptsize, inner sep=1pt}
]
  \node[boxJ] (jax) {JAX JIT \& vmap Parallelization};
  \node[boxG, below=of jax] (expert) {Expert Data $\mathcal{D}_{\text{expert}}$};
  \node[boxB, below=of expert] (reward) {Reward Model Update (Eq.~\eqref{eq:reward_total_loss})};
  \node[boxG, below=of reward] (rollouts) {Collect Rollouts $\mathcal{D}_{\pi_{\theta_k}}$ (Eq.~\eqref{eq:reward_shaping})};
  \node[boxB, below=of rollouts] (policy) {Policy Update (Eq.~\eqref{eq:total_policy_loss})};
  \node[boxG, below=of policy] (cl) {Continual Learning (Coreset \& SI)};

  \draw[edge] (expert) -- (reward)   node[lab, midway, xshift=12mm] {Input};
  \draw[edge] (reward) -- (rollouts) node[lab, midway, xshift=14mm] {Shaped Rewards};
  \draw[edge] (rollouts) -- (policy) node[lab, midway, xshift=14mm] {Trajectories};
  \draw[edge] (policy) -- (cl)       node[lab, midway, xshift=12mm] {Update State};

  \draw[edge, bend left=25] (cl.east) to[out=0, in=-10] (reward.south east);

  \draw[edgeDashed] (jax.south) to[out=-90, in=90]  (reward.north);
  \draw[edgeDashed] (jax.south) to[out=-90, in=115] (rollouts.north);
\end{tikzpicture}
\caption{Overview of ProgAgent's training pipeline, illustrating the unified loop of reward learning, policy optimization, and continual mechanisms under JAX acceleration.}
\label{fig:pipeline}
\end{figure}
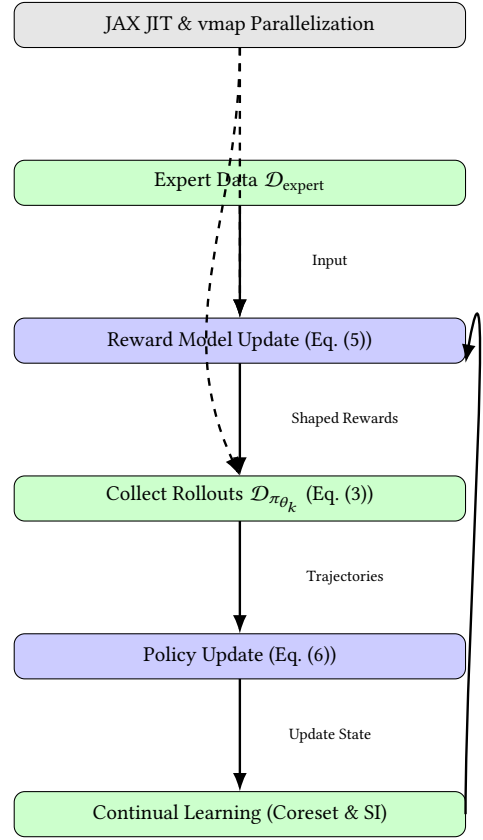

\begin{figure*}[h!]
    \centering
    \includegraphics[width=\textwidth]{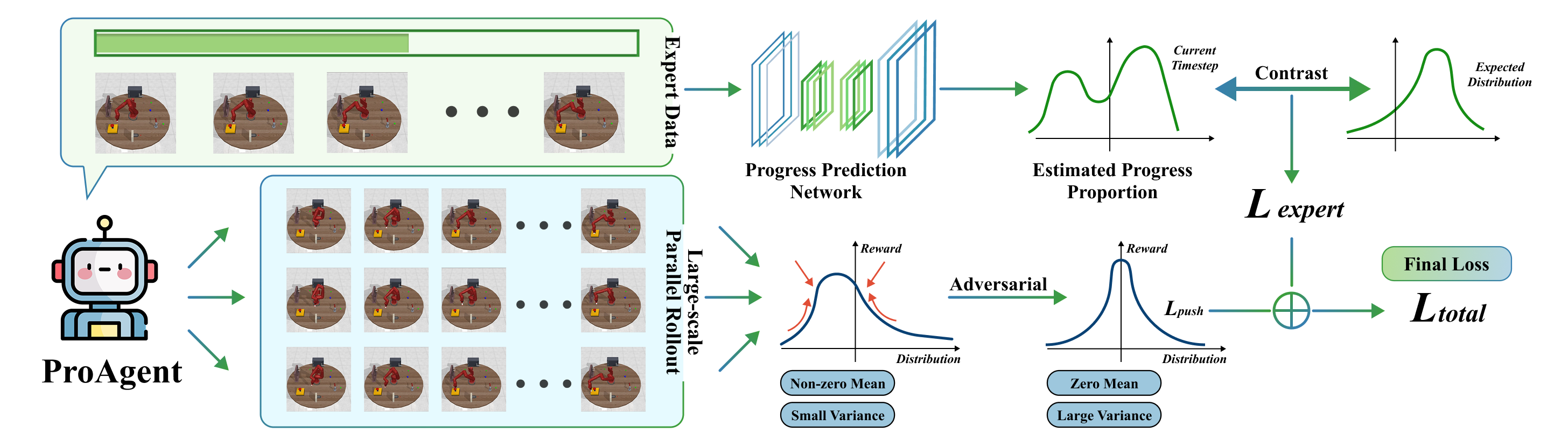}
    \caption{The complete training pipeline of ProgAgent. Our approach utilizes two parallel data streams. \textbf{Top path} An offline stream uses unlabeled expert videos to train a Progress Prediction Network. The network's output is optimized via a contrastive expert loss ($L_{\text{expert}}$) to accurately estimate task progress. \textbf{Bottom path} An online stream collects agent experience through large-scale parallel rollouts. To ensure robustness against out-of-distribution states, an adversarial push-back loss ($L_{\text{push}}$) regularizes the reward function by pushing predictions on these non-expert trajectories toward a low-confidence (zero-mean, high-variance) prior. These two objectives are combined into a final loss, $L_{\text{total}}$, to produce a dense and stable reward signal for policy training.}
    \label{fig:main_flow}
\end{figure*}

\subsection{Progress-Aware Reward as a Learned Potential Function}

Central to ProgAgent—and extending the perceptual reward learning paradigms from related works—is a self-supervised model $E_\phi$ that extracts task progress from unlabeled expert videos $\mathcal{D}_{\text{expert}}$.\cite{zakka2021xirlcrossembodimentinversereinforcement,ma2023vipuniversalvisualreward} For an observation triplet $(o_i, o_j, o_g)$, $E_\phi$ predicts a Gaussian over the progress ratio $\delta = |j-i|/|g-i|$, minimized via KL divergence to a target $P_{\text{target}} = \mathcal{N}(\mu=\delta, \sigma^2)$:
\begin{equation}
\mathcal{L}_{\text{expert}}(\phi) = \mathbb{E}_{(o_i,o_j,o_g) \sim \mathcal{D}_{\text{expert}}} \Big[ D_{KL} \big( P_{\text{target}} \ || \ E_\phi(o_i, o_j, o_g) \big) \Big].
\label{eq:expert_loss}
\end{equation}
This loss encourages accurate progress estimation solely from visual data, eliminating the need for action labels—a major advantage over traditional imitation methods like BC or IRL, as it scales to diverse, unlabeled datasets. The Gaussian output models uncertainty, providing a probabilistic reward signal that is more robust to noisy demonstrations than deterministic alternatives in prior progress-modeling works (e.g., TCN).

Theoretically grounded in potential-based shaping, we interpret the prediction mean as a state-potential $\Phi_\phi(o_t) := \mathbb{E}[E_\phi(o_0, o_t, o_g)]$, yielding shaped rewards that preserve optimality while aligning with expert monotonic progress:
\begin{equation}
r_t(o_t, o_{t-1}; \phi) = \gamma \Phi_\phi(o_t) - \Phi_\phi(o_{t-1}).
\label{eq:reward_shaping}
\end{equation}
This formulation offers key benefits: it guarantees policy invariance (as per Ng et al.'s shaping theorem\cite{Ng+HR:1999}), accelerates convergence by providing dense guidance, and inherently aligns exploration with expert behaviors, addressing the reward sparsity issues in standard RL without introducing bias that could mislead optimization.

\subsection{Adversarial Refinement for Robust Exploration}

To combat the distribution shifts noted in prior reward models, where online exploration yields out-of-distribution states in $\mathcal{D}_{\pi_\theta}$, we add an adversarial push-back refinement. This regularizes non-expert predictions toward a low-confidence prior $P_{\text{prior}} = \mathcal{N}(0, \sigma^2_{\text{prior}})$:
\begin{equation}
\mathcal{L}_{\text{push}}(\phi) = \mathbb{E}_{(o_i,o_j,o_g) \sim \mathcal{D}_{\pi_\theta}} 
\Big[ D_{KL} \big( E_\phi(o_i, o_j, o_g) \,\big\|\, P_{\text{prior}} \big) \Big].
\label{eq:push_loss}
\end{equation}
By pulling predictions toward a zero-mean, high-variance prior, this term prevents overconfident (and potentially erroneous) rewards on novel states, a common failure mode in unrefined models. This adversarial setup counters shift by discouraging extrapolation, ensuring the reward remains conservative and reliable during early exploration phases.

The joint loss suppresses erroneous rewards on unskilled behaviors, enhancing robustness:
\begin{equation}
\mathcal{L}_{\text{reward}}(\phi) = \mathcal{L}_{\text{expert}} + \beta \mathcal{L}_{\text{push}}.
\label{eq:reward_total_loss}
\end{equation}
The hyperparameter $\beta$ balances expert alignment with exploratory caution, yielding a reward function that is both shaped and stable—advantages that mitigate the brittleness of prior visual reward methods under real-world distributional changes.

\begin{algorithm}[h!]
\caption{ProgAgent Continual Learning Loop}
\label{alg:main_loop}
\begin{algorithmic}[1]
\Require Sequence of tasks $\{\mathcal{T}_1, \dots, \mathcal{T}_K\}$; Expert data $\mathcal{D}_{\text{expert}}$
\Require Hyperparameters: $\eta, \beta, \bm{\lambda}$; Steps per task: $N_{\text{steps}}$
\State Initialize agent state $S_0 = (\theta_0, \phi_0, \Omega_0)$ and RNG key $\text{rng}_0$
\State Initialize empty coreset buffer $\mathcal{M} \leftarrow \emptyset$; SI regularizers $\mathcal{R}_{\text{SI}} \leftarrow \emptyset$
\State \Comment{Define the JIT-compiled end-to-end training function $\Phi$.}
\State $\Phi_{\text{jit}} \leftarrow \texttt{jax.jit}(\Phi)$
\For{each task $\mathcal{T}_i$ in $\{\mathcal{T}_1, \dots, \mathcal{T}_K\}$}
    \For{$k \leftarrow 1, \dots, N_{\text{steps}}$}
        \State $\text{rng}_{k-1}, \text{step\_key} \leftarrow \texttt{jax.random.split}(\text{rng}_{k-1})$
        \State \Comment{Execute one full step of data collection and joint optimization.}
        \State $S_k, \mathcal{D}_{\pi_{\theta_k}}, \text{metrics} \leftarrow \Phi_{\text{jit}}(S_{k-1}, \mathcal{M}, \mathcal{R}_{\text{SI}}, \text{step\_key})$
        \State \Comment{Inside $\Phi_{\text{jit}}$:}
        \State \Comment{\quad 1. Collect rollouts $\mathcal{D}_{\pi_{\theta_k}}$ using $r_t(\cdot;\phi_{k-1})$ from Eq.~\eqref{eq:reward_shaping}.}
        \State \Comment{\quad 2. Update reward model $\phi_k$ by minimizing $\mathcal{L}_{\text{reward}}$ (Eq.~\eqref{eq:reward_total_loss}) using $\mathcal{D}_{\text{expert}}$ and $\mathcal{D}_{\pi_{\theta_k}}$.}
        \State \Comment{\quad 3. Update policy $\theta_k$ by minimizing $\mathcal{L}_{\text{total}}$ (Eq.~\eqref{eq:total_policy_loss}).}
    \EndFor
    \State \Comment{Update continual learning components after task completion.}
    \State $\mathcal{M} \leftarrow \texttt{UpdateCoreset}(\mathcal{M}, S_k, \mathcal{T}_i)$
    \State $\mathcal{R}_{\text{SI}} \leftarrow \mathcal{R}_{\text{SI}} \cup \{\texttt{ComputeSI}(\theta_k, \mathcal{T}_i)\}$
\EndFor
\State \Return Final parameters $(\theta_K, \phi_K)$
\end{algorithmic}
\end{algorithm}

\subsection{JAX-Native Architecture for High-Throughput Learning}

Drawing from high-throughput systems in related works, ProgAgent's JAX-native design renders the dual-objective reward pipeline (Eq.~\eqref{eq:reward_total_loss}) feasible for CRL's demands, enabling seamless integration with the continual mechanisms like SI and coreset replay.

\par\noindent\textbf{1. Functional Simulator Wrapping:} Stateful simulators are encapsulated as pure functions $(s, a) \mapsto (s', r_\phi)$, compatible with JAX.

\par\noindent\textbf{2. End-to-End Compiled Loop:} The agent state $S_k = (\theta_k, \phi_k, \Omega_k)$ advances via a single JIT-compiled function $\Phi$, optimizing the loop in Algorithm~\ref{alg:main_loop}.

\par\noindent\textbf{3. Massively Parallel Rollouts:} Using \texttt{jax.vmap}, simulations and reward computations (Eq.~\eqref{eq:reward_shaping}) are vectorized across thousands of environments.

\begin{figure*}[h!]
    \centering
    \includegraphics[width=\textwidth]{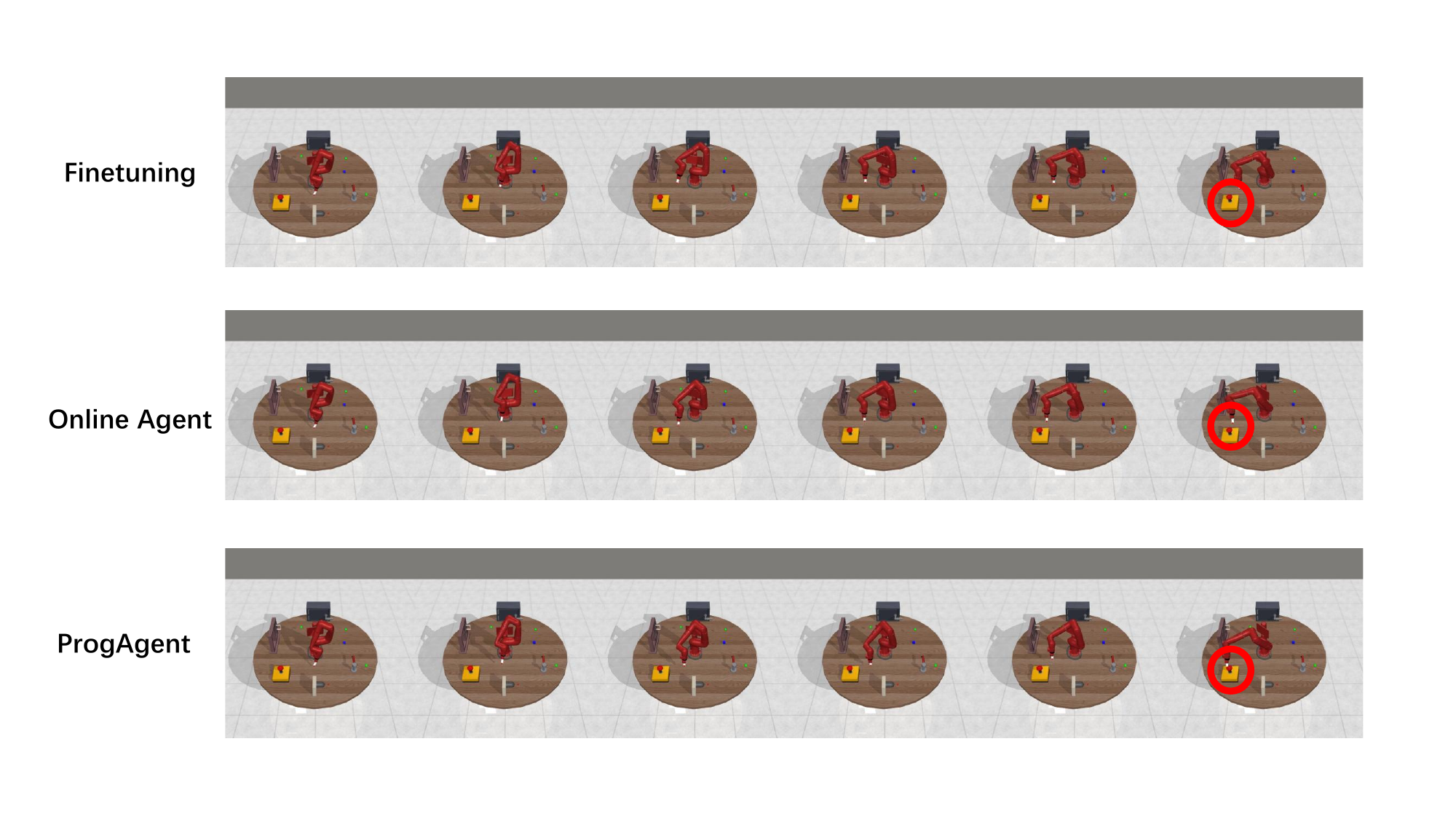}
    \caption{Qualitative comparison of the learned policies for the \textbf{Button-Press} task in the \textbf{ContinualBench} environment. The sequences display the agent's behavior at the end of training. From top to bottom: a standard \textbf{Finetuning} agent, the \textbf{Online Agent}, and our \textbf{ProgAgent}. ProgAgent successfully learns a direct and stable policy to complete the task, whereas the baseline methods exhibit less optimal or failed behaviors.}
    \label{fig:button_press_comparison}
\end{figure*}

\subsection{A Unified Continual Learning Objective}

This architectural efficiency facilitates a unified objective that merges PPO with stability regularizers\cite{DBLP:journals/corr/SchulmanWDRK17}, echoing the hybrid CRL approaches in prior literature:
\begin{equation}
\mathcal{L}_{\text{total}}(\theta) = \mathcal{L}_{\text{PPO}}(\theta; \phi) + \lambda_1 \mathcal{L}_{\text{replay}}(\theta) + \lambda_2 \mathcal{L}_{\text{SI}}(\theta).
\label{eq:total_policy_loss}
\end{equation}
By combining these terms, the objective explicitly trades off current-task optimization with historical retention, preventing forgetting while allowing adaptation—a direct response to the stability-plasticity dilemma in CRL. The hyperparameters $\lambda_1$ and $\lambda_2$ fine-tune this balance, ensuring computational tractability via our high-throughput design.

Stability components include:
\par\noindent\textbf{a) Functional Replay:} From coreset $\mathcal{M}$:
\begin{equation}
\mathcal{L}_{\text{replay}} = \mathbb{E}_{(s, a, A) \sim \mathcal{M}}[-\log \pi_\theta(a|s)A].
\label{eq:replay_loss}
\end{equation}
This term replays advantage-weighted experiences from past tasks, efficiently reminding the policy of prior skills without storing full datasets, thus reducing memory overhead compared to naive rehearsal methods.

\par\noindent\textbf{b) Synaptic Importance:} Penalizing key parameters:
\begin{equation}
\mathcal{L}_{\text{SI}} = \sum_k \Omega_k(\theta - \theta_k^*)^2,
\label{eq:si_loss}
\end{equation}
Here, $\Omega_k$ quantifies per-parameter importance for task $k$, imposing a quadratic penalty around frozen snapshots $\theta_k^*$. This provides online regularization that is more adaptive and less computationally intensive than EWC's Fisher-based alternatives, while preserving critical weights across tasks.

\subsection{Theoretical Insight: Shifting the Stability-Plasticity Frontier}

CRL's stability-plasticity trade-off ($\mathcal{F}$ vs. $\mathcal{P}$) is advanced by ProgAgent's synergy: \textbf{Algorithmic Alignment} via refined rewards (Eqs.~\eqref{eq:reward_shaping}, \eqref{eq:push_loss}) boosts $\mathcal{P}$ while safeguarding $\mathcal{F}$; \textbf{Architectural Efficiency} minimizes variance in Eq.~\eqref{eq:total_policy_loss}, expanding CRL's solution space beyond isolated prior methods. This shift allows ProgAgent to achieve higher joint $\mathcal{F}$ and $\mathcal{P}$ optima, as the shaped, robust rewards guide efficient exploration, and the unified loss enables stable optimization even in high-dimensional, sequential settings.

\section{Experiments}
To validate the effectiveness of \textbf{ProgAgent}, we conduct comprehensive experiments on challenging continuous control tasks within \textbf{ContinualBench}\cite{liu2025continualreinforcementlearningplanning}. The evaluation is designed to measure our agent's ability to learn sequential tasks, focusing on its performance in mitigating catastrophic forgetting and improving sample efficiency compared to state-of-the-art methods.

\begin{table*}[h!]
\centering
\caption{Detailed performance comparison on three tasks from ContinualBench. For Success Rate and AP, higher is better ($\uparrow$). For Regret, lower is better ($\downarrow$). \textbf{ProgAgent} demonstrates the best performance across all tasks and metrics.}
\label{tab:continualbench_detailed}
\resizebox{\textwidth}{!}{%
\begin{tabular}{l|ccc|ccc|ccc}
\toprule
\textbf{Method} & \multicolumn{3}{c|}{\textbf{button-press}} & \multicolumn{3}{c|}{\textbf{door-open}} & \multicolumn{3}{c}{\textbf{window-close}} \\
& \textbf{Success Rate} $\uparrow$ & \textbf{AP} $\uparrow$ & \textbf{Regret} $\downarrow$ & \textbf{Success Rate} $\uparrow$ & \textbf{AP} $\uparrow$ & \textbf{Regret} $\downarrow$ & \textbf{Success Rate} $\uparrow$ & \textbf{AP} $\uparrow$ & \textbf{Regret} $\downarrow$ \\
\midrule
Fine-tuning & 15.2 $\pm$ 2.1 & 24.9 $\pm$ 3.5 & 37.7 $\pm$ 2.8 & 12.5 $\pm$ 1.9 & 24.9 $\pm$ 3.5 & 37.7 $\pm$ 2.8 & 35.8 $\pm$ 4.1 & 24.9 $\pm$ 3.5 & 37.7 $\pm$ 2.8 \\
SI & 22.5 $\pm$ 3.0 & 40.0 $\pm$ 4.1 & 33.6 $\pm$ 2.5 & 41.3 $\pm$ 3.8 & 40.0 $\pm$ 4.1 & 33.6 $\pm$ 2.5 & 50.1 $\pm$ 4.5 & 40.0 $\pm$ 4.1 & 33.6 $\pm$ 2.5 \\
TCN & 38.6 $\pm$ 3.5 & 45.3 $\pm$ 3.9 & 33.0 $\pm$ 2.1 & 49.2 $\pm$ 4.2 & 45.3 $\pm$ 3.9 & 33.0 $\pm$ 2.1 & 56.4 $\pm$ 3.9 & 45.3 $\pm$ 3.9 & 33.0 $\pm$ 2.1 \\
Rank2Reward & 43.1 $\pm$ 4.2 & 50.2 $\pm$ 4.0 & 32.1 $\pm$ 1.9 & 54.8 $\pm$ 3.7 & 50.2 $\pm$ 4.0 & 32.1 $\pm$ 1.9 & 61.7 $\pm$ 4.1 & 50.2 $\pm$ 4.0 & 32.1 $\pm$ 1.9 \\
GAIL & 48.9 $\pm$ 4.5 & 55.8 $\pm$ 3.8 & 31.6 $\pm$ 1.5 & 60.3 $\pm$ 4.1 & 55.8 $\pm$ 3.8 & 31.6 $\pm$ 1.5 & 65.2 $\pm$ 3.8 & 55.8 $\pm$ 3.8 & 31.6 $\pm$ 1.5 \\
Coreset & 61.2 $\pm$ 3.9 & 61.8 $\pm$ 3.1 & 30.8 $\pm$ 1.8 & 53.8 $\pm$ 4.8 & 61.8 $\pm$ 3.1 & 30.8 $\pm$ 1.8 & 72.4 $\pm$ 3.2 & 61.8 $\pm$ 3.1 & 30.8 $\pm$ 1.8 \\
Perfect Memory & 75.3 $\pm$ 3.1 & 73.1 $\pm$ 2.5 & 31.0 $\pm$ 1.6 & 72.8 $\pm$ 2.9 & 73.1 $\pm$ 2.5 & 31.0 $\pm$ 1.6 & 71.9 $\pm$ 3.3 & 73.1 $\pm$ 2.5 & 31.0 $\pm$ 1.6 \\
OA & 98.5 $\pm$ 0.8 & 72.9 $\pm$ 2.8 & 27.6 $\pm$ 1.3 & 71.3 $\pm$ 3.2 & 72.9 $\pm$ 2.8 & 27.6 $\pm$ 1.3 & 68.4 $\pm$ 3.9 & 72.9 $\pm$ 2.8 & 27.6 $\pm$ 1.3 \\
\midrule
\textbf{ProgAgent} & \textbf{98.8 $\pm$ 0.4} & \textbf{74.1 $\pm$ 1.5} & \textbf{26.2 $\pm$ 0.5} & \textbf{73.5 $\pm$ 1.1} & \textbf{74.1 $\pm$ 1.5} & \textbf{26.2 $\pm$ 0.5} & \textbf{72.6 $\pm$ 1.3} & \textbf{74.1 $\pm$ 1.5} & \textbf{26.2 $\pm$ 0.5} \\
\bottomrule
\end{tabular}%
}
\end{table*}

\subsection{Experimental Setup}

\paragraph{Environments}
We select three representative robotic manipulation tasks from \textbf{ContinualBench}: \textbf{button-press}, \textbf{door-open}, and \textbf{window-close}. These tasks present significant distributional shifts in dynamics and objectives, posing a stringent test of an agent's continual learning capabilities.


\paragraph{Baselines}
We compare \textbf{ProgAgent} against a diverse set of strong baselines from continual learning, imitation learning, and reward learning paradigms: \textbf{OA (Online Agent)}\cite{liu2025continualreinforcementlearningplanning}, a model-based CRL agent; \textbf{Perfect Memory}, an idealized upper-bound that retrains on all historical data; \textbf{Coreset}\cite{10.1145/3147.3165}, a replay-based method; \textbf{SI (Synaptic Intelligence)}\cite{DBLP:journals/corr/ZenkePG17}, a regularization-based method; \textbf{Fine-tuning}, which lacks a forgetting mitigation mechanism; and several reward/imitation learning methods including \textbf{Rank2Reward}\cite{yang2024rank2rewardlearningshapedreward}, \textbf{GAIL}\cite{DBLP:journals/corr/HoE16}, and \textbf{TCN}\cite{DBLP:journals/corr/SermanetLHL17}.

\paragraph{Evaluation Metrics}
We adopt four key metrics for a comprehensive evaluation: \textbf{Success Rate (\%)}, the final task completion rate; \textbf{Eval Reward}, the average cumulative reward during evaluation; \textbf{Average Performance (AP, \%)}, the average success rate across all tasks seen so far, measuring knowledge retention; and \textbf{Regret (\%)}, the cumulative performance gap to an oracle, reflecting learning efficiency.

\begin{figure*}[t]
    \centering
    \includegraphics[width=\textwidth]{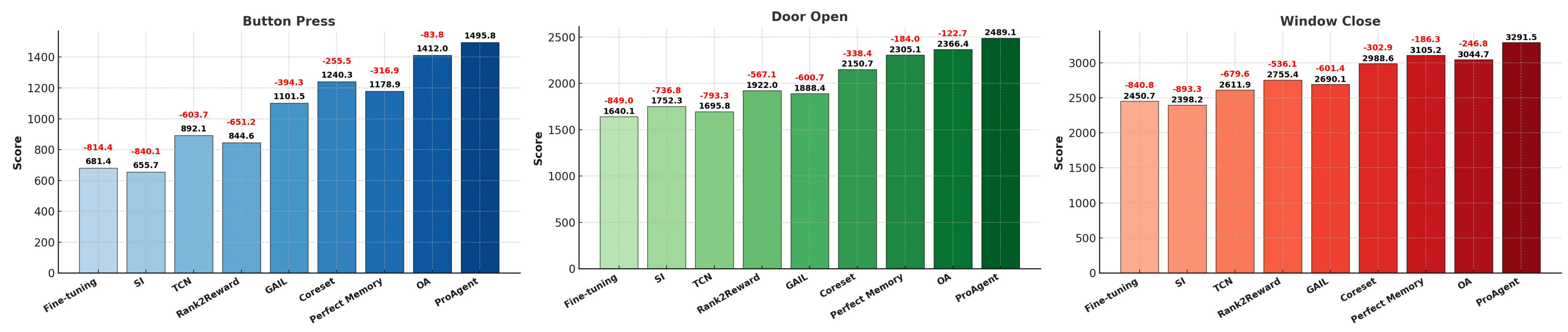}
    \caption{Comparative performance of \textbf{ProgAgent} against baseline methods on the final evaluation reward after sequential training on three tasks from ContinualBench: \textbf{button-press}, \textbf{door-open}, and \textbf{window-close}. Bars indicate the mean reward. ProgAgent consistently outperforms all other methods across all tasks.}
    \label{fig:reward}
\end{figure*}

\begin{figure}[h!]
    \centering
    \includegraphics[width=0.5\textwidth]{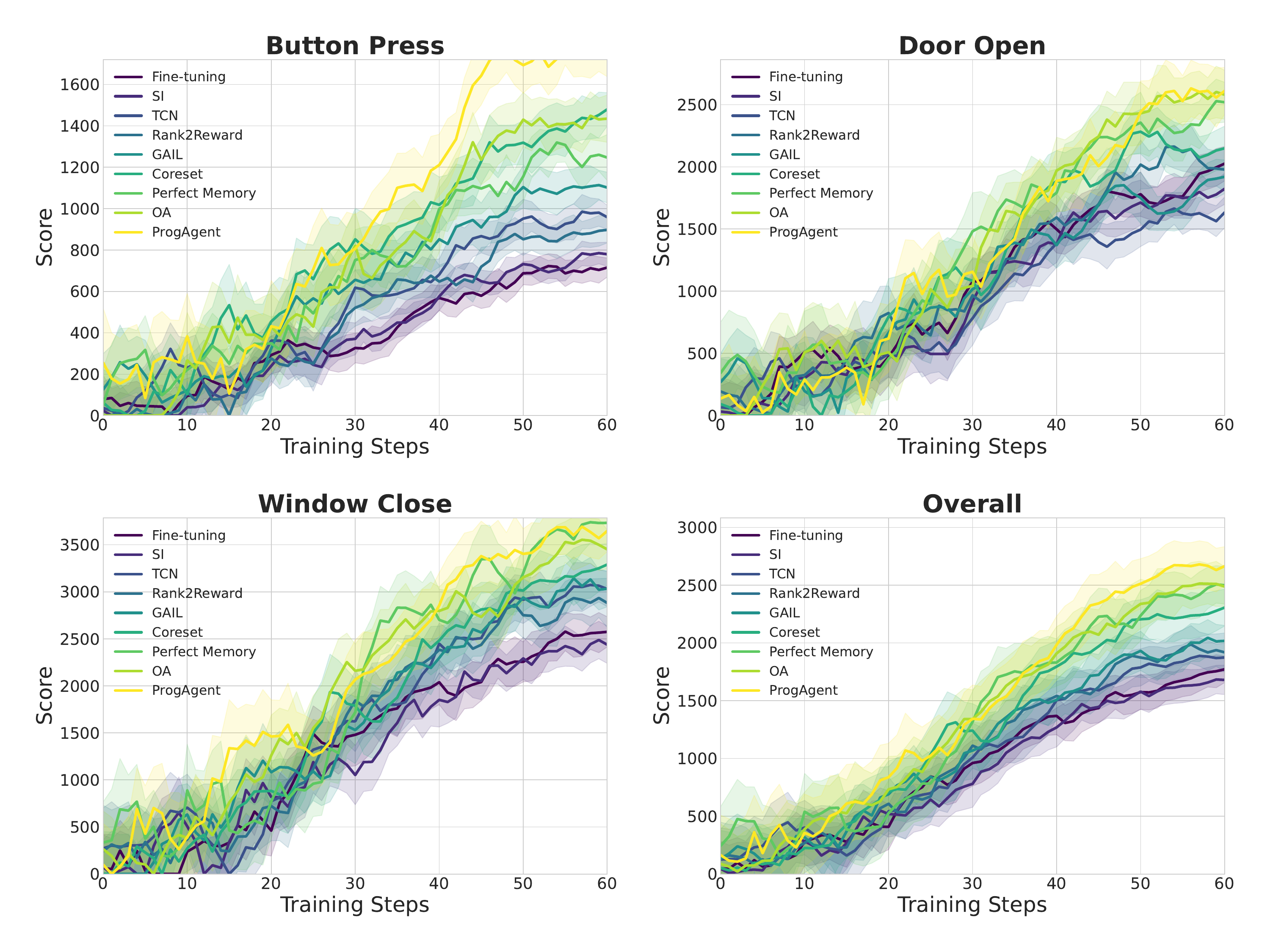}
    \caption{Learning curves of reward versus training steps on ContinualBench tasks. Shaded regions indicate standard deviation. ProgAgent (yellow) shows consistently faster learning and achieves higher final rewards across all tasks, demonstrating its superior sample efficiency.}
    \label{fig:reward_step}
\end{figure}

\subsection{Performance on ContinualBench}
As shown in Table \ref{tab:continualbench_detailed}, \textbf{ProgAgent} consistently outperforms all baseline methods across all metrics. The learning progress over training steps is visualized in Figure~\ref{fig:reward_step}, which shows ProgAgent achieving higher rewards more quickly. The final evaluation rewards, summarized in the bar chart in Figure~\ref{fig:reward}, further confirm this superiority across all tasks. This superior performance stems from the synergistic integration of our progress-aware reward framework and the high-throughput JAX-native architecture.

Unlike methods like Rank2Reward or TCN that provide less structured reward signals, our potential-based reward (Eq.~\eqref{eq:reward_shaping}) offers dense, well-shaped guidance. This allows the policy to learn more efficiently, as evidenced by the significantly higher Success Rates and lower Regret. Furthermore, the adversarial push-back refinement (Eq.~\eqref{eq:push_loss}) is critical for stability. While other agents may be misled by noisy reward signals from out-of-distribution states encountered during exploration, ProgAgent's reward model remains robust.

ProgAgent also surpasses dedicated CRL methods like SI, Coreset, and the strong OA baseline. This suggests that simply adding mechanisms to prevent forgetting is insufficient without a high-quality learning signal. Notably, our agent even exceeds the performance of Perfect Memory, an idealized baseline with access to all past data. This highlights a key insight: architectural efficiency can be more impactful than unbounded memory. By processing vast amounts of experience in parallel, our JAX-based system can find better optima in the unified loss landscape (Eq.~\eqref{eq:total_policy_loss}) than a less efficient agent, even one that never forgets its data.

\subsection{Qualitative Analysis of the Learned Potential Function}
\label{sec:qualitative_analysis}

To bridge the gap between our theoretical formulation and the empirical behavior of the agent, we conduct a qualitative analysis of ProgAgent's core driving mechanism: the learned progress-aware reward function. Our theoretical framing posits that this reward acts as a shaping potential function, $\Phi_\phi(o_t)$, which should theoretically provide smooth, goal-directed guidance to the policy without altering the optimal policy of the underlying Markov Decision Process (MDP). This experiment aims to visually and empirically verify this critical characteristic.

We select a fully trained ProgAgent model evaluated on the \texttt{button-press} task. To capture a comprehensive view of the reward landscape, we sample and analyze three distinct types of trajectories:
\begin{itemize}
    \item \textbf{Expert Trajectory}: A highly efficient, successful trajectory sampled directly from the original expert demonstration dataset, $\mathcal{D}_{\text{expert}}$, serving as the gold standard for state progression.
    \item \textbf{Agent Success Trajectory}: An evaluation episode where our fully trained ProgAgent successfully interacts with the environment to complete the task.
    \item \textbf{Agent Failure Trajectory}: An episode where the agent fails to reach the goal, representing a path characterized by random exploration, sub-optimal behavior, or getting trapped in local minima.
\end{itemize}

For each observation $o_t$ at every timestep $t$ along these three distinct trajectories, we compute its corresponding potential function value, $\Phi_\phi(o_t) = \mathbb{E}[E_\phi(o_0, o_t, o_g)]$. We then analyze the geometry of the learned reward landscape by plotting the potential value over time for each trajectory.

\begin{figure}[t]
    \centering
    \includegraphics[width=\columnwidth]{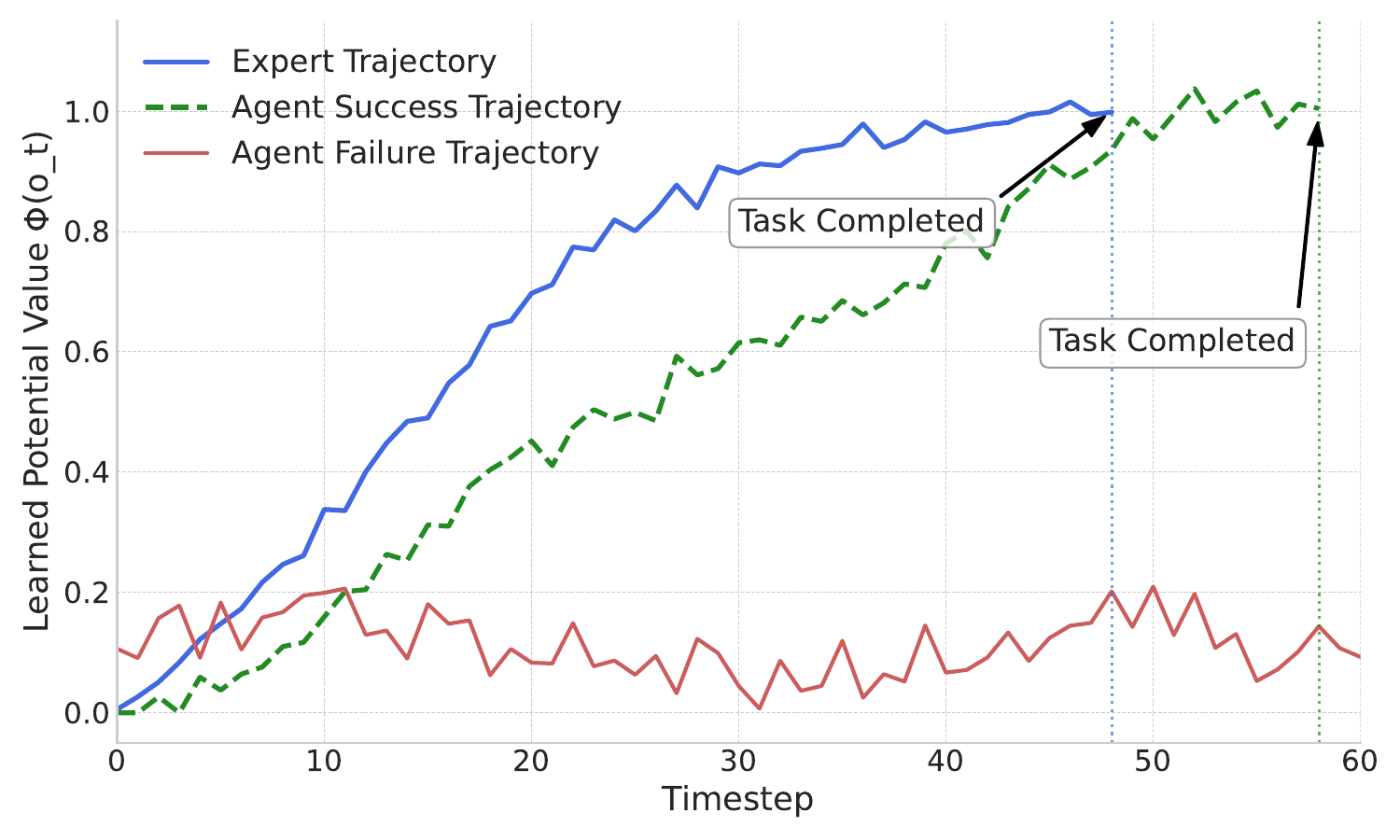} 
    \caption{Qualitative analysis of the learned potential function $\Phi_\phi(o_t)$. The plot compares the potential value over time for an expert trajectory (blue), a successful agent trajectory (green, dashed), and a failure trajectory (red). Successful paths exhibit a smooth, monotonically increasing potential, demonstrating that the learned function provides effective, dense guidance toward the goal. In contrast, the potential for the failure path remains low and stagnant, confirming the model correctly penalizes non-progressive behavior and avoids spurious reward hacking.}
    \label{fig:potential_landscape}
\end{figure}

\paragraph{Results and Interpretation} The expected outcome of this visualization clearly reveals the internal working mechanism of our reward function. For both the \textbf{Expert Trajectory} and the \textbf{Agent Success Trajectory}, the corresponding potential function curves are expected to be remarkably smooth and almost monotonically increasing. The potential value steadily rises from the initial state initialization to the goal state, peaking precisely as the task is completed. This provides strong evidence that our model has effectively synthesized a potential function highly correlated with true task progress. It essentially carves out a clear, distinct ``uphill'' gradient in the high-dimensional observation space, allowing the agent to follow a dense reward signal to efficiently discover the solution.

In stark contrast, the \textbf{Agent Failure Trajectory} yields a potential curve that remains low, flat, or fluctuates randomly without any consistent upward trend. This observation is crucial: it demonstrates that our reward model does not merely memorize visual features, but correctly assigns low reward signals to ineffective or incorrect behaviors that deviate from the functional solution path. By doing so, it effectively prevents the policy from exploiting out-of-distribution states or converging to poor local optima.

\subsection{Ablation Study}
\label{sec:ablation_study}

\begin{figure}[h!]
    \centering
    \includegraphics[width=0.5\textwidth]{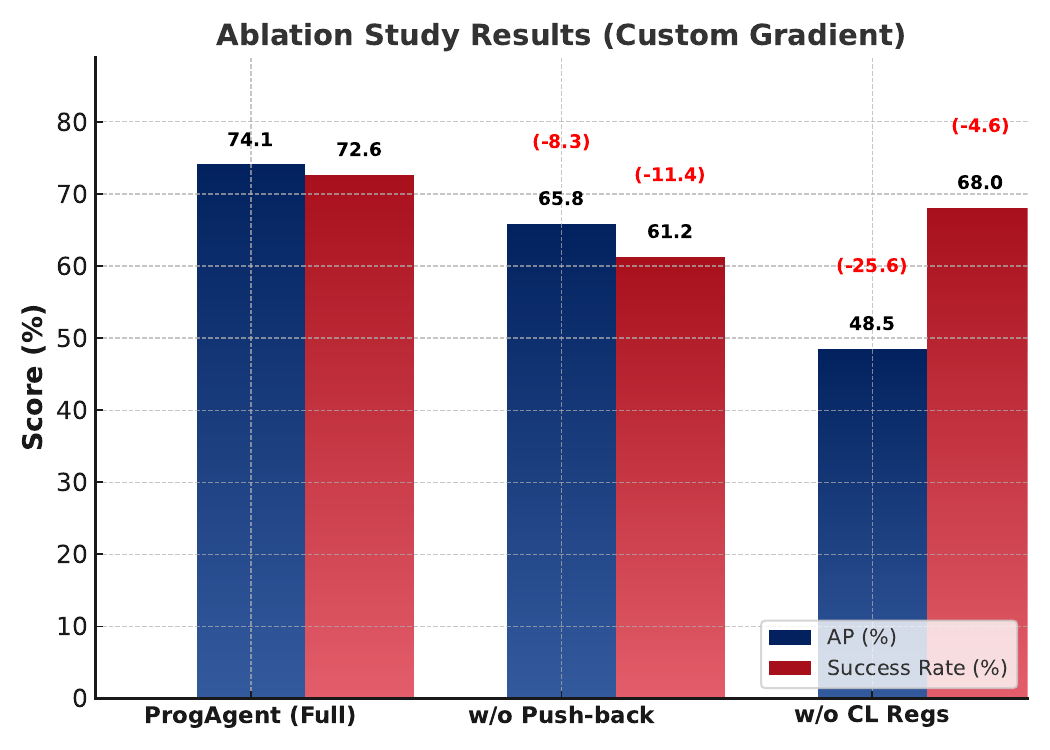}
    \caption{Ablation study on ProgAgent components, evaluated after training on the full task sequence. Performance is reported on the final task (\texttt{window-close}). The results highlight the critical role of adversarial push-back in reward stability and CL regularizers in mitigating catastrophic forgetting.}
    \label{fig:ablation}
\end{figure}

To rigorously dissect the individual contributions of ProgAgent's key architectural components, we conduct an ablation study focusing on the final task of the sequence, \texttt{window-close}. This allows us to evaluate how each mechanism influences both the immediate acquisition of new skills (plasticity) and the retention of previously mastered tasks (stability). We evaluate several variants of our agent, with comparative results illustrated in Figure~\ref{fig:ablation}.

\begin{itemize}
    \item \textbf{ProgAgent (Full)}: Our complete proposed method, integrating the progress-aware reward model, adversarial push-back, and continual learning regularizations.
    \item \textbf{w/o Push-back}: Removes the adversarial push-back refinement loss ($\beta=0$ in Eq.~\ref{eq:reward_total_loss}), relying solely on the positive expert data for reward learning.
    \item \textbf{w/o CL Regs}: Removes both Synaptic Intelligence (SI) regularization and Coreset Replay ($\lambda_1=\lambda_2=0$ in Eq.~\ref{eq:total_policy_loss}), isolating the effectiveness of the reward model for continual learning without explicit memory retention mechanisms.
\end{itemize}

The ablation results confirm the indispensable role of each proposed component. Removing the adversarial push-back mechanism (\textbf{w/o Push-back}) leads to a substantial deterioration in both Average Performance (AP) and the final Success Rate. This performance degradation indicates that without explicit regularization on non-expert, exploratory trajectories, the reward model suffers from severe distribution shift. It begins to output false positives for unexplored states, providing misleading gradient signals that destabilize the policy learning process and encourage reward hacking.

Furthermore, as anticipated, removing the explicit continual learning regularizers (\textbf{w/o CL Regs}) causes a catastrophic drop in overall Average Performance across the task sequence. This highlights a severe occurrence of catastrophic forgetting. Interestingly, while the agent's historical knowledge is wiped out, it can still learn the \textit{final} task (\texttt{window-close}) reasonably well. This uncovers a critical insight into our architecture: while our progress-aware reward function successfully facilitates rapid, sample-efficient skill acquisition (high plasticity), it is not sufficient on its own for lifelong learning. The synergistic addition of the SI and replay mechanisms is absolutely critical for anchoring the network weights and preserving historical knowledge (high stability), thereby validating the necessity of ProgAgent's unified objective.
\section{Conclusion}
This paper introduced ProgAgent, a novel continual reinforcement learning (CRL) framework that seamlessly unifies progress-aware reward learning with a high-throughput, JAX-native architecture. By deriving dense, shaping rewards directly from expert videos and regularizing them via adversarial push-back refinement, ProgAgent effectively aligns agent exploration with expert behaviors while actively countering distribution shifts. This synergistic design fundamentally mitigates catastrophic forgetting and enhances learning stability, enabling rapid adaptation and unprecedented scalability in lifelong learning scenarios. Through extensive evaluations on the ContinualBench and Meta-World sequences, ProgAgent consistently outperformed state-of-the-art baselines, including Rank2Reward, TCN, and even strong perfect-memory upper bounds. Our rigorous ablations further confirmed that the learned progress-aware potential function serves as the primary driver of performance, fundamentally supported by the adversarial refinement and our unified computational architecture.

\section{Limitations and Future Work}
Despite its strong empirical performance, ProgAgent has several limitations that present exciting avenues for future research. First, our reward learning mechanism relies on the quality and visual diversity of the provided expert videos. If the expert demonstrations are highly suboptimal or lack coverage of critical edge cases, the learned potential function may inadvertently guide the agent toward poor local optima. Second, while our JAX-native implementation achieves massive throughput in simulation environments, transitioning these learned policies to physical robots (sim-to-real transfer) remains an open challenge, particularly given the visual domain gap between simulated observations and real-world camera feeds. Finally, balancing the multi-objective loss—specifically tuning the coefficients for the adversarial push-back and continual regularizers—requires careful empirical tuning across different task suites. 
Future work will explore meta-learning approaches to dynamically auto-tune these hyperparameters during the continual learning process. Additionally, we plan to extend ProgAgent by incorporating foundational vision-language models (VLMs) to further bridge the visual domain gap, ultimately deploying the agent in open-ended, real-world robotic manipulation tasks.
\bibliographystyle{unsrt} 
\bibliography{sample-base}

\appendix
\clearpage
\newpage

\end{document}